\documentclass{egpubl}
\usepackage{gch2021}
 
\WsPaper           

\usepackage[T1]{fontenc}
\usepackage{dfadobe}  
\usepackage{amsmath}
\usepackage{amssymb}
\usepackage{booktabs}
\usepackage{caption}
\usepackage{subcaption}
\usepackage{cite}  
\BibtexOrBiblatex
\electronicVersion
\PrintedOrElectronic
\ifpdf \usepackage[pdftex]{graphicx} \pdfcompresslevel=9
\else \usepackage[dvips]{graphicx} \fi

\usepackage{egweblnk}

\DeclareMathOperator*{\argmin}{arg\,min} 

\newcommand{\etal}{\textit{et al.}}

\title[Riedones3D]{Riedones3D: a celtic coin dataset for registration and fine-grained clustering}

\author[S. Horache \& J.-E. Deschaud \& F. Goulette \& K. Gruel \& T. Lejars]{\parbox{\textwidth}{\centering S. Horache$^{1}$\orcid{0000-0002-5884-8667},J-E. Deschaud$^1$\orcid{0000-0002-6696-9354},  F. Goulette$^1$, K. Gruel$^{2}$, T. Lejars$^{2}$ and O. Masson$^{3}$}
        \\
{\parbox{\textwidth}{\centering $^1$MINES ParisTech, PSL University, Centre for Robotics, 75006 Paris, France\\
         $^2$ ENS Ulm, PSL University, AOROC, 75005 Paris, France $^3$Ecole Pratique des Hautes Etudes (EPHE) - 4-14 Rue Ferrus, 75014 Paris (France)}}
}

\begin{document}


\maketitle
\begin{abstract}
Clustering coins with respect to their die is an important component of numismatic research and crucial for understanding the economic history of tribes (especially when literary production does not exist, in celtic culture). It is a very hard task that requires a lot of times and expertise. To cluster thousands of coins, automatic methods are becoming necessary. 
Nevertheless, public datasets for coin die clustering evaluation are too rare, though they are very important for the development of new methods. 
Therefore, we propose a new 3D dataset of 2 070 scans of coins. With this dataset, we propose two benchmarks, one for point cloud registration, essential for coin die recognition, and a benchmark of coin die clustering. 
We show how we automatically cluster coins to help experts, and perform a preliminary evaluation for these two tasks.
The code of the baseline and the dataset will be publicly available at \url{https://www.npm3d.fr/coins-riedones3d} and \url{https://www.chronocarto.eu/spip.php?article84&lang=fr}.


\end{abstract}  

\section{Introduction}

Coins are a testimony of old civilizations, and their study is crucial for understanding economic history, especially when literary production is not consistent. In this regard, die study is very important for numismatic research~\cite{gruel2020barbus, gruel1981celt, callatay1995calc}. A die is a stamp that allows an image to be impressed upon a piece of metal for making a coin. Usually, the matrix of the die is bigger than the coin itself. 
Coin die clustering can greatly help specialists: it will allow the estimation of the number of coins that were emitted at a larger scale (Esty explains the process to estimate the number of produced coins from the clustering of dies~\cite{esty1986estim}). Coin die clustering will also allow for establishing the chronology of the production of these coins. However, coin die recognition is difficult and it demands a lot of times and great expertise for experts in numismatics. 
The most ambitious die study to date is from Wolfgang Fischer-Bossert and is based on 8 000 coins (silver didrachms, 510---280 BC.) and it took him around 10 years to complete the study~\cite{wolfgang1999}. There exist some hoards with more than 100 000 coins to study. 
Moreover, it requires a trained eye to recognize the dies, because of rust, wear or damage. Some coins are too worn, so it is not possible to perceive the differences with other coins. Because die recognition takes a lot of time and requires expertise, automatic methods are all the more necessary, especially to scale-up on the size of the study: it would allow for the processing of coins from museums that have not been analysed yet.

\begin{figure}[ht]
    \centering
    \includegraphics[width=\linewidth]{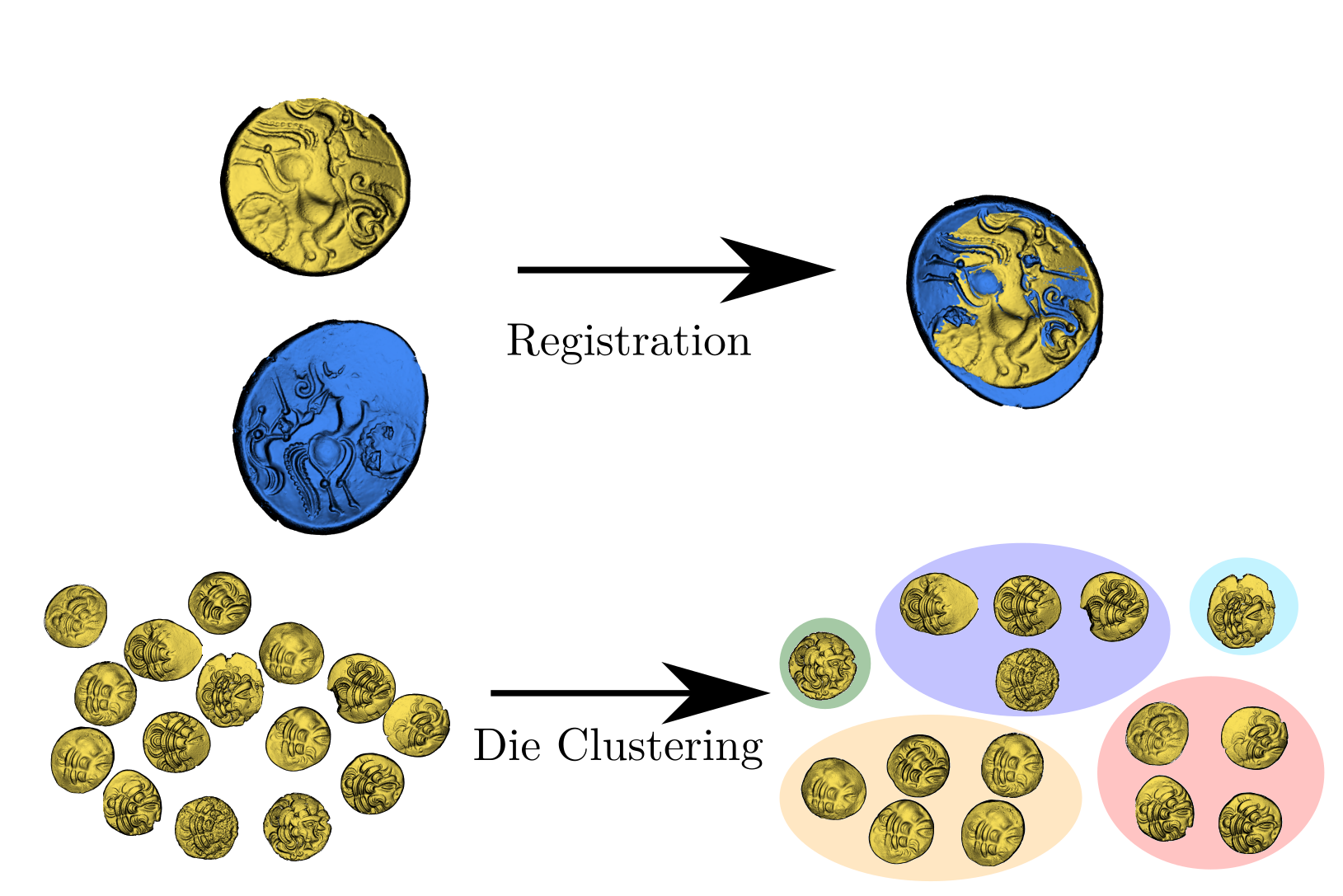}
    \caption{Proposed tasks on Riedones3D dataset: rigid registration between two coins and die clustering}
    \label{fig:catcheye}
\end{figure}

Some works have tried to automate this work using photos, but the results were not satisfying. The problem is also computationally expensive, because for $x$ coins, algorithms need to compare every pairs so there are $\frac{x(x-1)}{2}$ comparisons; for 1000 coins, there are around 500 000 comparisons to do. 
According to \cite{arandjelovic2020image}, "Die matching is an unexplored challenge in the realm of automated ancient coin analysis." It can be explained by the fact that public datasets of coins labeled by die are too rare.  Some coin datasets exist that are based on photos~\cite{arandjelovic2020image, Anwar2021CoinNetDA, zambanini_coarse_fine_2013}. However, none of them are labeled by die.
Contrary to high-level coin recognition, die matching requires details of the coin. 
That is why, in this paper, we propose a new dataset of 3D scans of celtic coins. 
Indeed, 3D scans from a structured light scanner can capture the fine geometry of the pattern.
Moreover, photos are altered by rust or lighting conditions~\cite{marchand2014gch.20141318}, whereas 3D scanners can capture only the geometry of the pattern.
Not only will this dataset be useful for experts in order to study coins with high-quality virtual data, but it is also an interesting dataset for the computer vision community in order to develop powerful new algorithms for very fine-grained pattern clustering. 

We propose a comparison of different registration algorithms between coins on this dataset (see Figure~\ref{fig:catcheye}). Registration of the pattern of two coins is a difficult problem but a necessary step for die clustering. Many elements in the coin, such as edges or cracks, perturb the registration. We will also provide a baseline for die clustering on this dataset (see Figure \ref{fig:catcheye}). Coin die clustering is challenging because the number of classes can be arbitrary and highly irregular. It is similar to a fine-grained clustering.


Our contributions are:
\begin{itemize}
    \item A curated dataset of high-quality 3D scans of coins labeled following their die by experts in numismatics.
    \item Strong baselines for registration and fine-grained clustering on that dataset.
\end{itemize}

\section{Related works and datasets}
\subsection{Coins recognition}
According to \cite{arandjelovic2020image}, the challenges of computer vision applied to numismatics are not correctly addressed. Indeed, most of the works of computer vision on numismatics are in a controlled environment and cannot be applied to real-world conditions. We think that it is because public datasets with labeled data are still rare in this domain, and the code is not always open-source.
Because data acquisition takes time and 3D scanners are expensive, few researchers work on 3D data \cite{TOLKSDORF2017400, Zambanini2009HistoricalCI}. Their datasets contain only dozens of coins.
Hence, there are few methods for analysing 3D data.
Because it is easier to get images, most of the methods in the literature uses images\cite{salgado_medieval_2016, schlag_ancient_2017, zambanini_coarse_fine_2013, zambanini_improving_2013, Anwar2021CoinNetDA}, and datasets can be up to 24 000 coins \cite{Anwar2021CoinNetDA}.  There are datasets from museum websites or auction websites \cite{schlag_ancient_2017, Anwar2021CoinNetDA, zambanini_coarse_fine_2013}. Nevertheless, these datasets are not labeled by die, because it requires a lot of expertise to label coins by die. Moreover, when coins are very worn, it is difficult to see correctly the details of the patterns just with photos, even though such pattern details are important for coin die recognition.
 Hence, most of these coin photo datasets are not made for coin die recognition. Usually, these datasets are used to classify high-level properties of the coins (which class, style, which emperor, etc.). The problem is different from ours. 
For coin die recognition, datasets and methods are very rare~\cite{arandjelovic2020image, }. Moreover, most of the methods are semi-automatic~\cite{TOLKSDORF2017400, lista2019simple}. 
For example, \cite{TOLKSDORF2017400} worked on coin die study for Roman coins. They have also scanned coins with a high-quality scanner. However, they only worked with 37 coins. 
\cite{horache2020auto} propose an automatic method for coin die recognition for 3D data, and showing promising results. It can be used as a baseline for our dataset.
Our dataset consists of 2 070 high-quality 3D scans of coins labeled by die, and is public.

\subsection{Datasets for pattern recognition}
For pattern clustering, metric learning (or similarity learning) plays an important role.
Metric learning can be used for pattern clustering, but datasets for metric learning have different applications. The applications of metric learning are usually face recognition, object retrieval or few-shot learning, so many datasets exist for these tasks, especially 2D image datasets.  
In 3D, there are datasets from SHREC challenges (SCHREC 2018 and SCHREC 2021 for cultural heritage). Especially SHREC 2021 dataset~\cite{shrec2021} is based on 3D textured models of objects (jar, bowls, figurine, etc.) and the challenges are to find similar objects by shape or by culture. However, it is different from the Riedones3D dataset. The Riedones3D dataset and SHREC2021 have the same dataset size but Riedones3D has much more classes, and the classes are highly unbalanced. Moreover, Riedones3D is a dataset of coins with fine-grained patterns (dies), which is very different from bowls or jars.

\subsection{Datasets for point cloud registration}
As Horache \etal~\cite{horache2020auto} highlight, 3D coin clustering can be decomposed into two steps: a registration step to align patterns, then, pair similarity computation in order to cluster data. Therefore, we propose to evaluate point cloud registration on Riedones3D because it is an important step for pattern clustering.     
Point cloud registration is applied in many domains, so a lot of datasets are available. We can divide registration datasets into two types: 
\begin{itemize}
    \item object-centric
    \item indoor and outdoor scene
\end{itemize}
For object-centric, the most famous dataset is The Stanford 3D scanning repository\cite{curless1996bunny}, with the bunny, widely used for experiments in the Computer Graphics community. However, the main problem is that the dataset is rather small for evaluation. 
Researchers add synthetic noise or outliers to test their registration algorithm, but it is not necessarily realistic.
Recently, a lot of deep learning methods have emerged to improve point cloud registration. Stanford 3D scanning repository is too small for training deep learning methods, therefore researchers use ModelNet~\cite{Wu_2015_CVPR} even though ModelNet was designed for point cloud classification. ModelNet is a synthetic dataset of CAD objects: it contains 40 classes and 12~311 shapes. Because it is a synthetic dataset, it is not representative enough of real-world data. The geometry is too simple in comparison with the fine patterns of the coins of Riedones3D. And these benchmarks contains only small-size point clouds. So some methods developed on these datasets do not work on real world datasets such as scans of indoor scenes (see experiment section of~\cite{choy2020deep}).

Due to the increasing number of 3D scanners, a lot of different datasets exist for example, indoor scans of RGB-D frames (3DMatch\cite{zeng20163dmatch}, TUM\cite{sturm12iros}) or outdoor scenes acquired by a LiDAR such as ETH~\cite{Pomerleau:2012}, KITTI~\cite{Geiger2012CVPR} or WHU-TLS~\cite{DONG2017whutls}. 
These real-world datasets are huge, have point clouds with massive numbers of points, and are more challenging than object-centric datasets. Usually, registration algorithm are used for 3D reconstruction or SLAM. However, these datasets are very different from datasets such as Riedones3D. They don't have the same challenges for registration. Coins are rather flat structures, and coins have elements such as cracks or edges that can perturb the registration.

\section{Construction of Riedones3D}
\begin{figure}
     \centering
     \includegraphics[width=\linewidth]{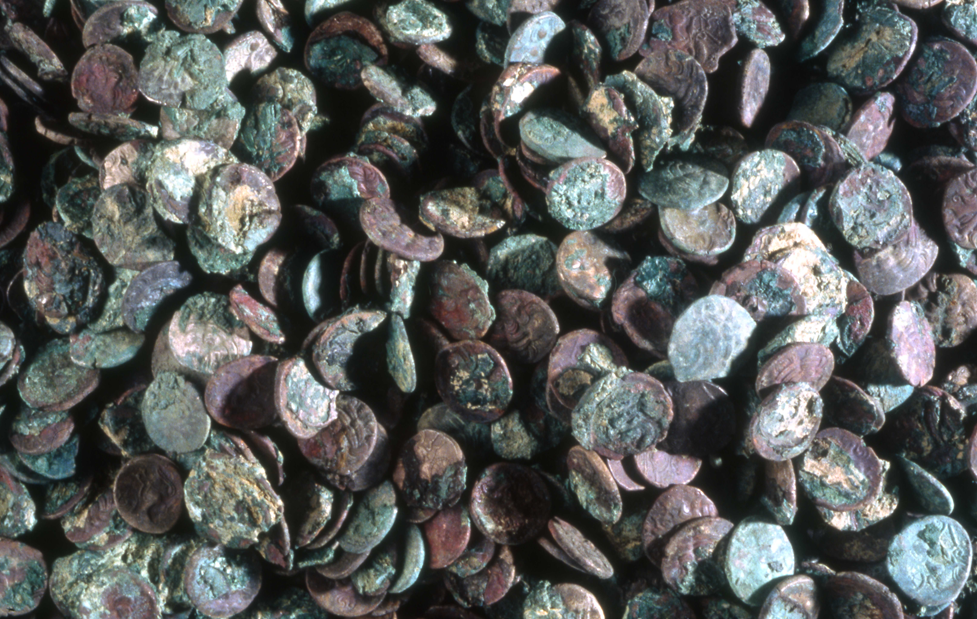}
     \caption{The Riedones hoard. We can see that coins are not necessarily in good condition, and that is why 3D scans help a lot.}
     \label{fig:treasure}
 \end{figure}

The Riedones were a Gallic tribe from the Armorican Peninsula (modern Brittany). Little is known about the Riedones: there are too few written sources. In the book Commentarii de Bello Gallico (commentaries on the gallic War) of Julius Ceasar, the Riedones are briefly mentioned. Around the city of Rennes, many Riedones artifacts have been found especially hoards. The coins from the proposed dataset were found in the town of Liffré (near the city of Rennes) and are conserved in the Museum of Brittany in Rennes. This hoard (see Figure~\ref{fig:treasure}) is exceptional, probably because these coins have not circulated so much, this is why, in this hoard, a lot of coins have the same die.

\begin{figure*}
    \centering
    \includegraphics[width=\textwidth]{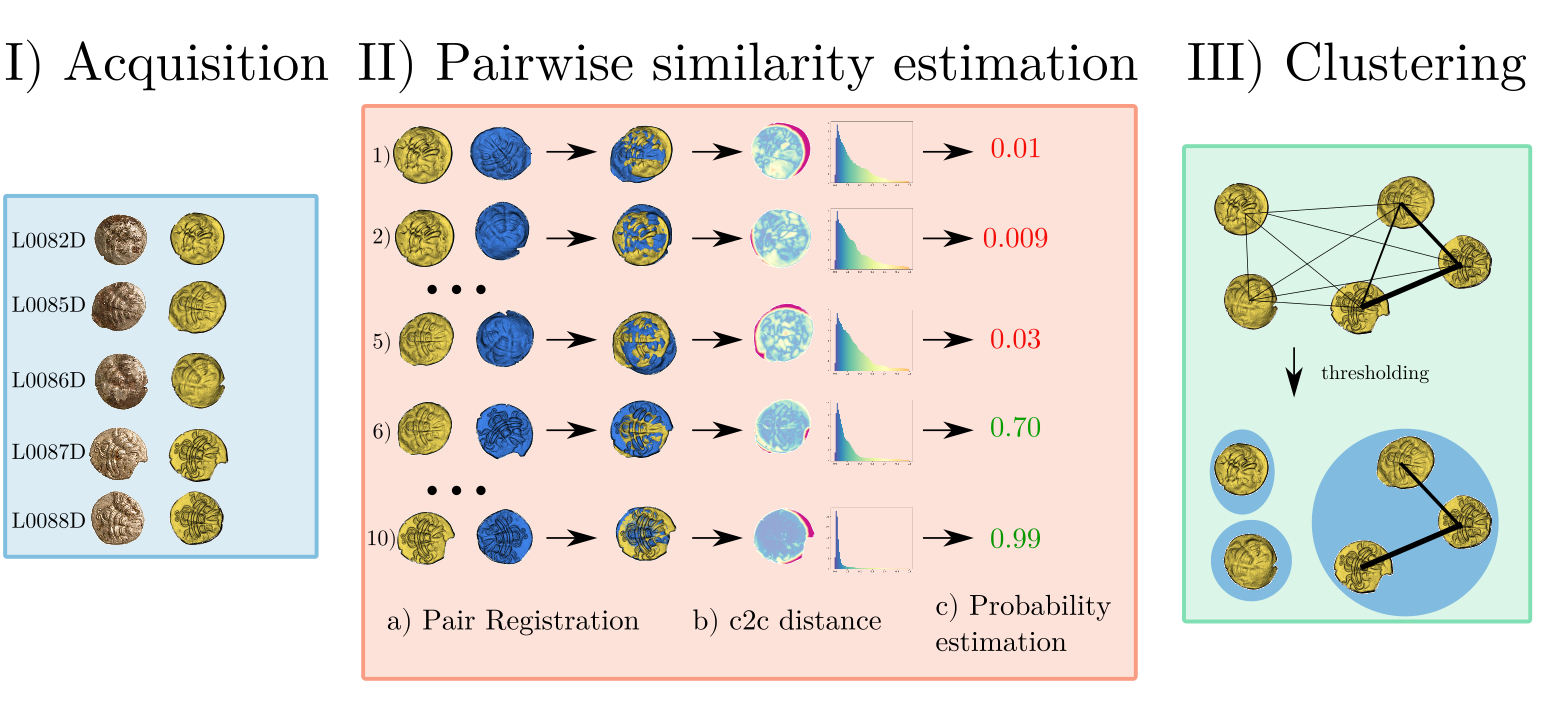}
    \caption{Summary of the proposed method. I) Data acquisition with a 3D scanner (in blue). II) Pairwise similarity estimation (in red): the goal is to estimate the probability that two coins were struck by the same die. First, we align patterns using a registration algorithm. Then, we compute the cloud-to -cloud distance (c2c distance) and its histogram. Then, we estimate the probability that two coins are from the same die using logistic regression. For the color of the probabilities, green is that the coins are from the same die (red is the contrary). III) Clustering (in green): with the graph of pairwise similarity, we can apply a threshold to remove low probabilities (represented by thin links). Clusters are connected components.}
    \label{fig:pipeline}
\end{figure*}

In order to annotate the coins quickly, we built a tool that will automatically pre-label coins based on an improvement of the method of~\cite{horache2020auto}. We bring some significant modifications: notably, we show that deep learning can be effective for the registration step.
The whole process can be summarized into three steps (see Figure~\ref{fig:pipeline}):
\begin{enumerate}
    \item data acquisition
    \item automatic pairwise similarity estimation
    \item automatic clustering and manual correction
\end{enumerate}
The following subsections will detail each step.

\subsection{Data acquisition}
\begin{figure}
     \centering
     \includegraphics[width=\linewidth]{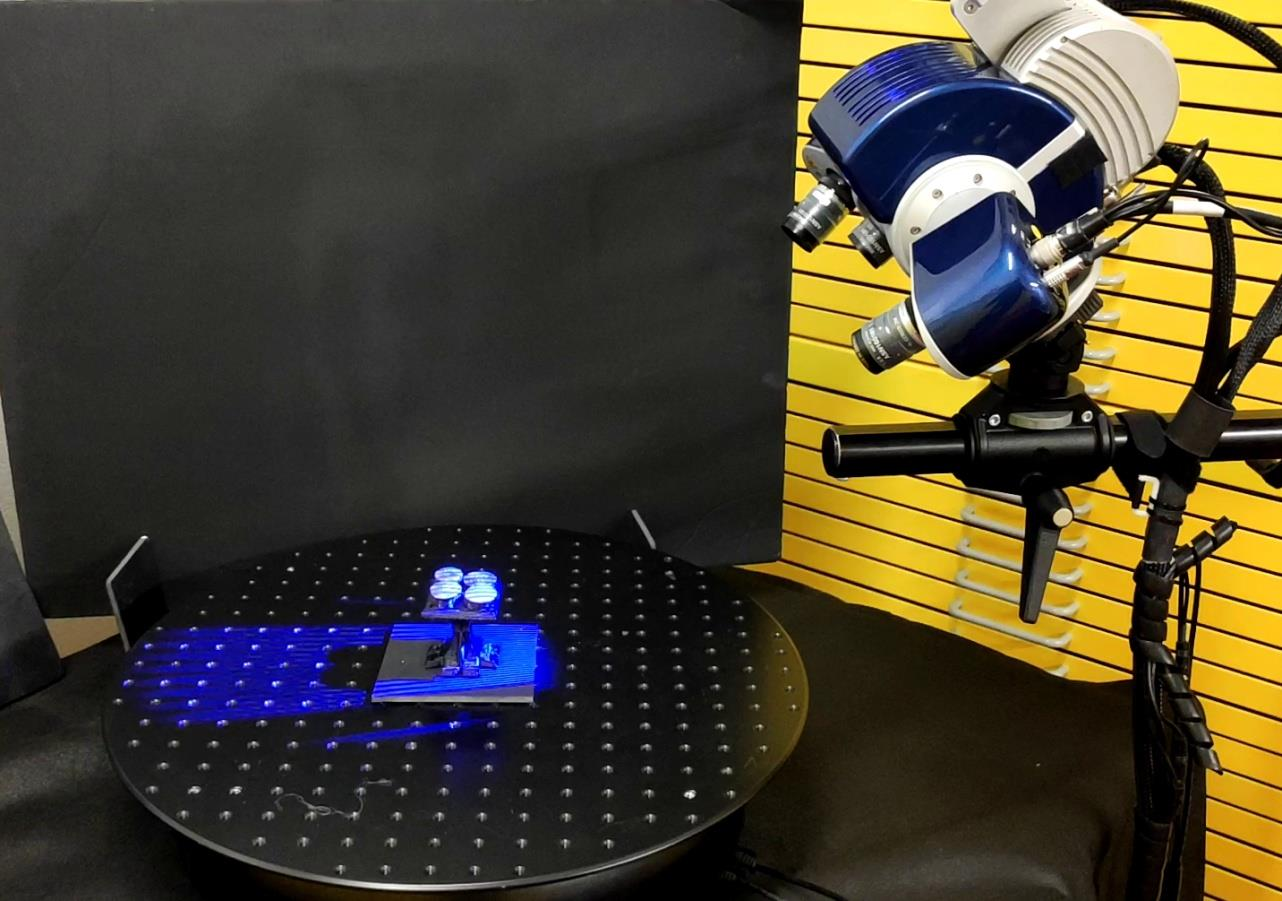}
     \caption{Photo of the acquisition process. The scanner is a SmartScan AICON, which uses structured blue light in order to acquire 3D geometry.}
     \label{fig:scanner}
 \end{figure}

\begin{figure}
    \centering
    \includegraphics[width=\linewidth]{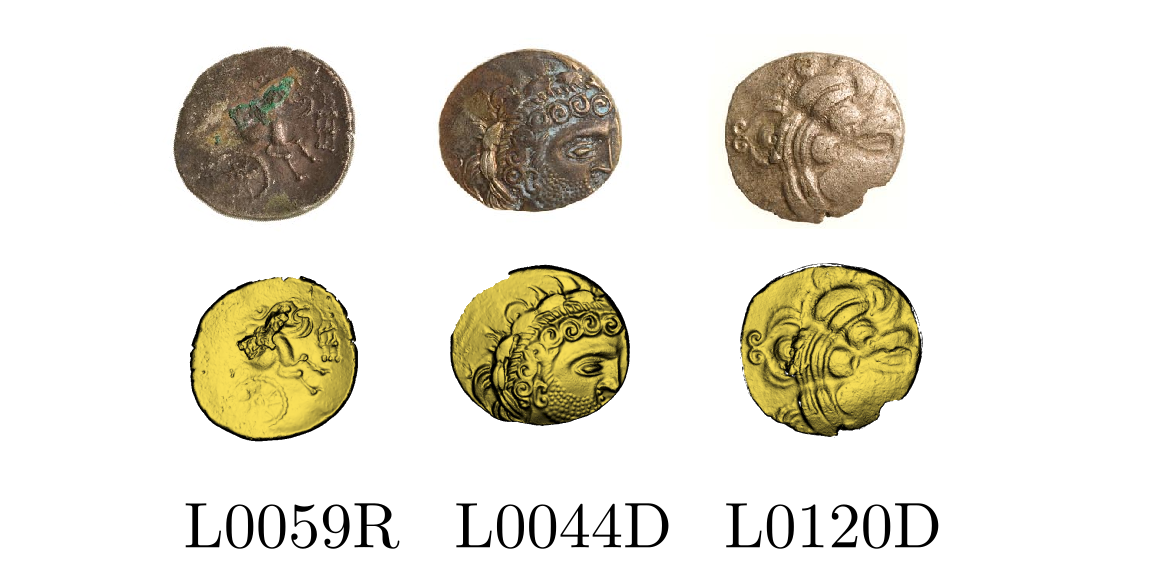}
    \caption{Top, photos of Riedones coins. Bottom, 3D scans of the same coins. From left to right, an example of obverse with beard, obverse without beard, and reverse.}
    \label{fig:example}
\end{figure}

Contrary to photos, 3D scans from high-quality scanners can highlight very fine details of the pattern, which can be crucial for die studies. For a coin of 2 cm diameter, a resolution has to be an order of magnitude of $10^{-1}$ mm in order to perceive some important details. Moreover, with photos, it is very hard to see the geometry of a pattern. We can see photos of coins in Figure~\ref{fig:example}: some patterns are stained by rust, and are not very visible. Also, 3D data can be used by experts to reconstruct a virtual version of the die and manipulate it virtually. 
That is why we decided to use a high-quality scanner (see Figure~\ref{fig:scanner}). 
 The coins have been acquired using AICON SmartScan. This scanner can scan objects with a resolution of 0.04 mm. Moreover, we need to scan the two faces of the coin: the obverse and the reverse. To be more efficient, we scanned coins 4 by 4 (see Figure~\ref{fig:scanner}). With the scanner software Optocat, some pre-processing has been done (hole filling, better alignment, outlier removal). Then, we obtained a dense mesh of the coin. We used the vertices of the mesh as our input point cloud and kept the normals per point (computed from the mesh). To scan and pre-process 4 coins, so 8 faces (obverses and reverses), it takes around 10 minutes.
 
\subsection{Pairwise similarity estimation}
In order to cluster coins by die, we need to compute a similarity between coins. To compute a similarity between a pair, we first align the patterns using a registration algorithm. Then, we compute the cloud-to-cloud distance and next, the probability that the coins were struck by the same die using the histogram of distances (see also Figure~\ref{fig:pipeline}).

\subsubsection{Point Cloud Registration}
In order to know whether two coins were struck by the same die, we need to align the coins. If we can align patterns at least partially, it means that the patterns come from the same die.
Rigid registration is the task of finding the rotation and translation that best align two point clouds (in our case, patterns of coins). 
Let $X=\{x_1, x_2 \dots x_n\}$ and $Y=\{y_1, y_2\dots y_m\}$ be point clouds represented by a set of 3D points.
Mathematically, registration can be described as:
\begin{align}
    (R^*, t^*, M^*) = \argmin_{R \in SO(3), t \in \mathbb{R}^3, M \in \mathcal{M}} \sum_{(i, j) \in M} ||Rx_i + t - y_j||^2
\end{align}
$SO(3)$ is the set of rotations and $\mathcal{M}$ is the set of set of matches.
If we know the correct matches, computing the rotation and translation is possible using the Kabsch algorithm. However, we do not know which matches are correct. 
A well-known algorithm for point cloud registration is Iterative Closest Point (ICP)~\cite{Besl1992AMF}. ICP~\cite{Besl1992AMF} has two steps:
\begin{enumerate}
    \item estimate matches searching for the closest point
    \item use the Kabsch algorithm to estimate the transformation
\end{enumerate} 
These two steps are repeated iteratively. In the case of coins, ICP~\cite{Besl1992AMF} has two major drawbacks. First, it is a local algorithm which means that using the closest point as a match is a good heuristic only when we are not too far from the right solution. Secondly, even if we are close to the right solution, we will have a lot of false matches because of some points on cracks or edges.
In other words, the edges or cracks can perturb ICP.

\paragraph{Point cloud registration using deep learning}
\begin{figure*}
    \centering
    \includegraphics[width=\textwidth]{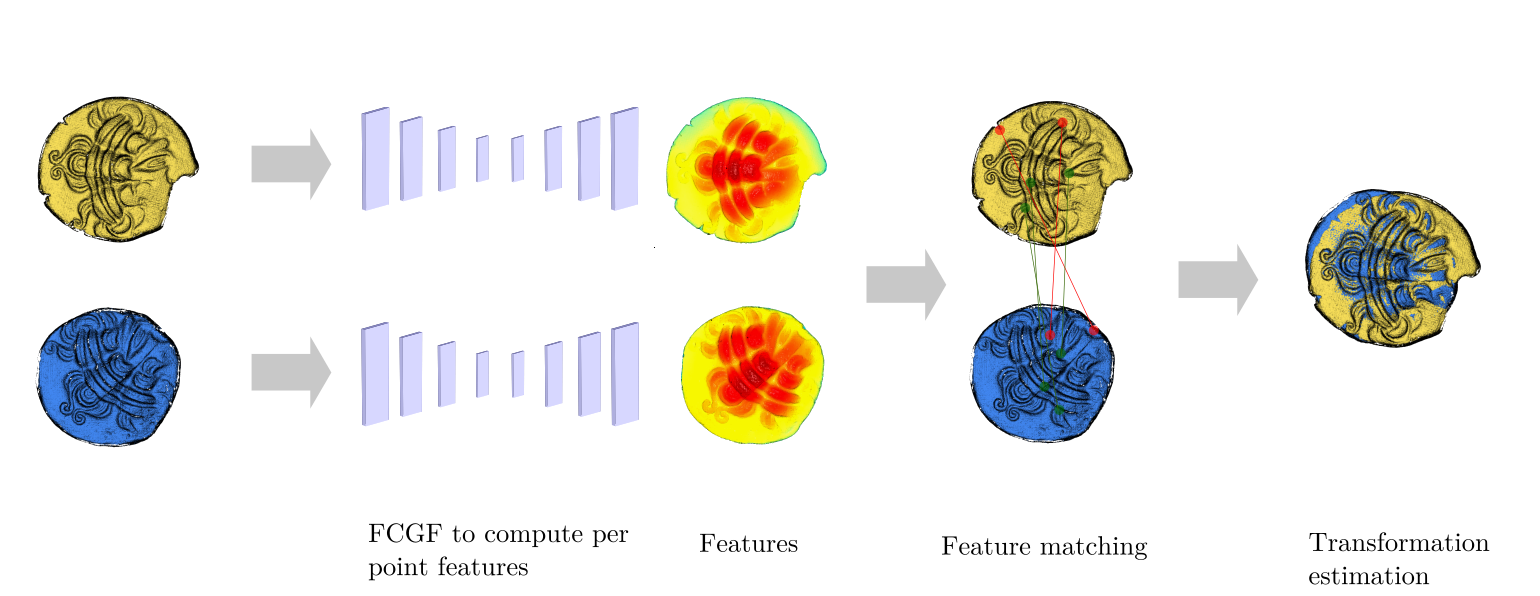}
    \caption{Registration using deep learning. First, we compute features using FCGF~\cite{Choy_2019_ICCV}. Then, we compute matches with the features. Finally, we use TEASER, a robust algorithm to estimate the transformation. Features are visualized with colors using Principal Component Analysis.}
    \label{fig:deep}
\end{figure*}

To solve the problems above, we can use registration with descriptor matching and especially we use deep learning (see Figure \ref{fig:deep}) to compute descriptors. We use Fully Convolutional Geometric Feature (FCGF~\cite{Choy_2019_ICCV}) to compute descriptors: it is a deep learning method that compute descriptors per point. FCGF uses a U-Net architecture \cite{Ronneberger2015unet} (widely used in 3D semantic segmentation~\cite{thomas2019KPConv, choy20194d, qi2017pointnet}), which is composed of an encoder and a decoder. FCGF allows to compute compact descriptors of dimension 32. To deal with large point clouds, FCGF uses sparse convolutions~\cite{choy20194d}. Therefore, FCGF is adapted to the Riedones3D dataset.
In order to train descriptors, FCGF uses a contrastive loss $L$ with hard negative sampling.

\begin{align}
    L =  &\sum_{(i, j) \in M^+} \{ [\|F_{X_i} - F_{Y_j}\| - m_+]^2_{+} \\
    & + \frac{1}{2} [m_{-} - \min_{k|(i, k) \in M^-} \|F_{X_i} - F_{Y_k}\|]^2_{+} \\
    & + \frac{1}{2} [m_{-} - \min_{k|(k, j) \in M^-} \|F_{X_k} - F_{Y_j}\|]^2_{+} \}
\end{align}
where $[.]_{+} = \max(., 0)$,   $M^+$ is the set of positive matches, and $M^-$ is the set of negative matches. $F_X = \{F_{X_1} \dots F_{X_n}\}$ (rest $F_Y$) is the set of descriptors computed on $X$ (resp. $Y$) using FCGF. $m_{+}$ and $m_{-}$ are hyper parameters of the contrastive loss. The goal is to minimize $L$ with respect to the parameters of the neural network.

Intuitively, descriptors computed by FCGF will have a small euclidean distance when it corresponds to the same part of the pattern. Thus, FCGF allows to compute descriptors invariant to rigid transformation, some small deformations, cracks and wear. However, training FCGF requires annotated data.

\paragraph{Manual registration and labeling of 200 coins}
To train FCGF, we need pairs of coins with positive matches (we can sample negative matches on the fly). To get positive matches, we computed ground truth transformations by manually picking pairs of points between pairs of coins. 200 coins were registered manually (only obverses without beard). It took two months to do the manual registration of these 200 coins and label them by their die. We trained FCGF on these coins using the ground truth. We found that FCGF can generalize on different obverse patterns (with and without beard) but it can also generalize on reverse patterns (with a horse and a wheel). In other words, we found that registration using FCGF works well on different patterns, even if the training set is small and not diverse. Training FCGF on Riedones3D tooks 4 days with an Nvidia RTX 1080Ti.

\paragraph{Descriptor matching and robust estimation}
With FCGF, we have a descriptor for each point of the point cloud. 
We then take a random number $n$ of points (with their descriptors) on the two point clouds and try to find the matches between them for the computation of the rigid transformation. In our experiments, we tested with $n=250$ and $n=5000$.
For each descriptor $F_{X_i}  \in F_X$, we searched the closest descriptor in $F_Y$; we performed the opposite with the closest searched descriptor and kept only symmetric matches. 
But this method of matching does not guarantee that the match will be correct. Outliers can be numerous. Therefore, we cannot directly use the Kabsch algorithm and we need a robust estimator instead.
RANSAC~\cite{ransac} and TEASER algorithms~\cite{yang2020teaser}  are adapted for a robust estimation of the transformation. We used TEASER in our experiments because it is as good as RANSAC but faster.
Also, after estimating the transformation robustly, we can apply an ICP on point clouds around the positive matches in order to refine the registation.

\subsubsection{Pairwise similarity}
With the methods described above, we can correctly align the patterns of two coins. However, we need a method to measure similarity between aligned patterns. The proposed solution is to compute the cloud-to-cloud distance between point clouds of coins (c2c distance).

Suppose $X$ and $Y$ are aligned ($X$ is the source, and $Y$ is the target), the sample set of point-to-point distances $D =\{d_1, d_2 \dots d_m\}$ and $D' = \{d_1', d_2' \dots d_n'\}$ is defined as:
\begin{align}
    d_i = \min_{k = 1 \dots m}||x_i - y_k||, \quad d_i' = \min_{k = 1 \dots n}||x_k - y_i||
\end{align}
with $n$ the number of points in $X$, $m$ the number of points in $Y$, $t$ the number of samples taken from each point cloud.

Then we compute an histogram of $D$ and $D'$, and finally, we compute the mean of the two histograms. We discard distances that are too big. The final histogram will be the input of a logistic regression that will estimate the probability that the coins were struck by the same die. We used the 200 manually labeled obverses (explained above), in order to train the logistic regression. 
With this method, we can obtain good accuracy for the binary classification (two coins comes from the same die or not) and display good clustering results ($97\%$ accuracy for the obverses). However, there may still be errors. It took a few days to compute similarities between all pairs of coins for the whole Riedones3D dataset.

For each pair, the pairwise similarity estimation method (registration + c2c distance + probability estimation) takes 4.4 s (see Table~\ref{tab:time}).

\begin{table}[]
    \small
    \centering
    \begin{tabular}{l|ccc}
    \toprule
        & \textbf{Pair registration} & \textbf{Probability estimation} & \textbf{Total}\\
        \midrule
        Time (in s) &  3.8 & 0.6 & 4.4\\
    \bottomrule
    \end{tabular}
    \caption{Time (in s) for each operation of the pairwise similarity estimation pipeline (for one pair of scans). FCGF is computed on the GPU. The other operations are computed on the CPU. Each pair can be processed independently.}
    \label{tab:time}
\end{table}

\subsection{Clustering and Correction}

\subsubsection{Clustering}
Pairwise similarities can be represented as an undirected weighted complete graph, where each node represents a point cloud representing a face, and the link represents the probability that the pair of coins was struck by the same die. To obtain clusters, we remove links that have a probability below a threshold $\tau$ and then we compute the clusters by connected components. The choice of $\tau$ is difficult: if $\tau$ is too low, we obtain false positives (two coins not from the same die in the same cluster); if $\tau$ is too high, we remove true positives (coins from the same die not in the same cluster). Therefore, the choice of this parameter depends on the data: we will show how to fix it (from a small-size dataset labeled by an expert). It is also handy for the expert to try out different parameters, so that he can check manually which pairs are close.

\subsubsection{Manual correction}

An expert is necessary to check whether the coins have been well clustered or not. Sometimes, the machine can make errors because the pattern is worn out or the coin is bent. In that case, the expert must verify whether the clustering is correct quickly. Hopefully, verification is easier. First, thanks to registration, we don't need much expertise to see whether two coins come from the same die or not. Pair verification is faster with registration, because we can quickly see whether the pattern is aligned or not, and whether the algorithm made a mistake. 

\begin{figure}
    \centering
    \includegraphics[width=0.9\linewidth]{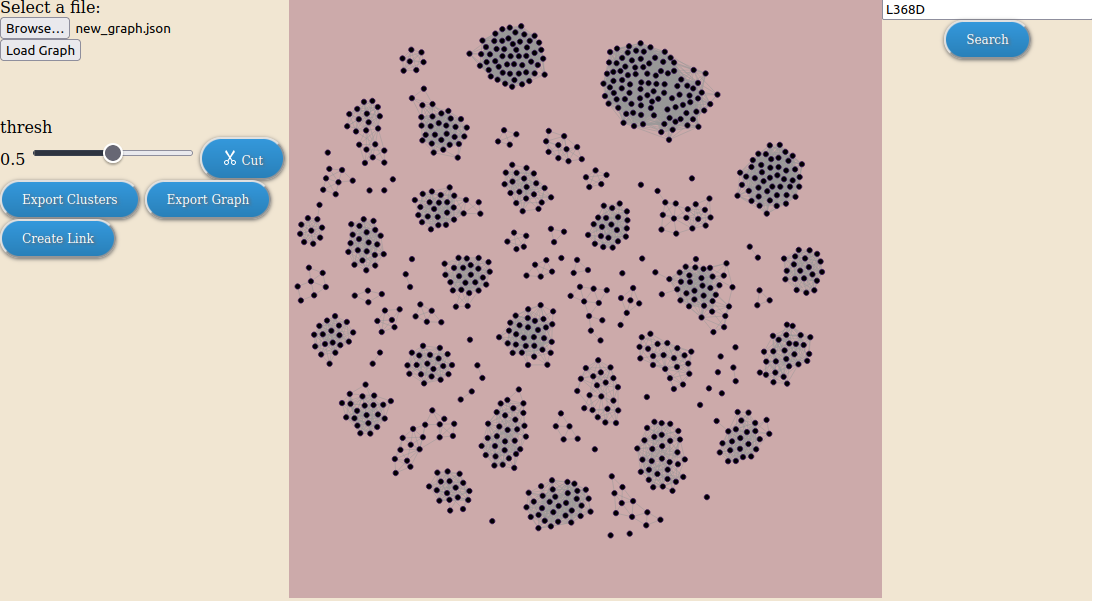}
    \caption{Interactive graph of similarities between coins (grouped by die). The user can quickly and seamlessly add new link or cut a wrong link. The user can also save the new graph and export the clusters.}
    \label{fig:graph}
\end{figure}

Moreover, we implemented an interactive graph of association. Each node represents a coin and each link weight represent the probability that two coins are from the same die (a screen capture is available in Figure \ref{fig:graph}). 
The graph is automatically obtained using the method above, but the user can  specify the threshold for the clustering.
Also, the expert can quickly search a coin and can quickly see the connections with the other coins.
This graph is a powerful tool for visualization, but also for edition. The expert can also edit the graph (add and remove links) and export the clusters.
Thus, verification is faster than with manual clustering, because we do not need to check every link. With this tool, it took few weeks to verify and correct the different clusters of coins for our whole dataset. We also corrected registration manually when it failed. 

\subsection{Implementation details}
For FCGF~\cite{Choy_2019_ICCV}, we use the implementation of Pytorch Point3D~\cite{chaton2020tp3d} 
(the sparse convolution is implemented using Minkowski Engine library\cite{choy20194d}). The sparse voxel size is set at 0.1 mm. To train FCGF, as it has been said earlier, we used 200 manually labeled obverses: it makes a training set of 2~132 pairs (only coins of the same die can be used for the training). We train the model during 380 epochs using a stochastic gradient descent with a learning rate of 0.1, a momentum of 0.8, a weight decay of 0.0001 and an exponential scheduler of 0.99 for the learning rate. Data augmentation is random rotation (between -25 and +25 degrees for the $x$ and $y$ axis and between -180 and +180 for the $z$ axis). We also performed a small random scale on point clouds (between 0.9 and 1.2). The input feature is ones, as in the original paper. For the loss, we use a contrastive loss with a positive margin $m_+=0.1$ and a negative margin $m_-=1.4$.
For the clustering part, we needed to compute a histogram of distances: we used 70 bins. To compute the histogram, we discarded points with distance above 0.6 mm. Before computing the point to point distance, we down-sample the source point cloud with a voxel size of 0.1~mm, and we down-sample the target point cloud with a voxel size of 0.05~mm.

\section{Baseline for automatic registration and clustering}

\subsection{Dataset description}
In total, there are 2070 scans: 968 obverses and 1102 reverses. For obverses, there are 887 obverses without beard, and 81 obverses with beard. We make a distinction between obverses with and without beard, because the classification of obverses with beard is obvious compared to obverses without beard: it does not need any expertise and it has been processed separately in this study.
Each scan corresponds to only one face (obverse or reverse) of a coin: it has a unique ID, which is LxD or LxR. L means Liffré (the place the treasure has been found), x is a number, and D means obverse, R reverse (for example L0001D or L0145R). The coins from L0001D to L0081D are beard, the others are not beard. Each file is represented by a point cloud in ply binary format which contains the point positions, the normals and colors (artificial colors computed from the normals for visualization). Normals are computed from the mesh. The original meshes are also available. 

These coins come from 293 dies: 81 obverses (74 obverses without beard, 7 obverses with beard) and 212 reverses.
For the reverses, there are more dies because the dies have worn out more quickly.
Each die has an ID with Rx for the reverse, Dx for the obverse without beard and DBx for the obverse with a beard (R11 means the die 11 of reverses). 

For the different tasks presented, the dataset has been split into training sets and test sets.

\subsection{Point cloud registration }

\subsubsection{Dataset}
We present a benchmark to compare different registration algorithms based on our dataset.
For a pair of coins, the goal is to find the right transformation with translation and rotation. As pointed out by~\cite{horache2020auto}, the problem is challenging because the coin edges perturb the registration. This kind of problem is rather rare in other registration datasets. Therefore, very few works are robust to this problem. In each die of the test set for registration, each pair is evaluated.

The registration benchmark is based on part of the Riedones3D dataset: the test set is composed of 160 obverse faces (80 faces without a beard and 80 faces with a beard) and 79 reverse faces. We use 27 dies in total for the evaluation. The benchmark has 2158 pairs with generated random rotations (for axis $x$ and $y$ between -25 and 25 degrees and for axis $z$ between -180 and 180 degrees). For every method, we downsample the point cloud with a grid size of 0.1 mm.
For deep learning methods, a training set of 200 obverse coins is available. We only train on the obverse without beard pattern to see if methods work on unseen obverse patterns but also if they generalize well on reverse patterns. 
For evaluation, we measure the Scaled Registration Error (SRE) as defined in~\cite{fontana2020benchmark}.
Let $X=\{x_1, \dots x_n \}$ and $Y=\{y_1, \dots y_m\}$ two point clouds. Let $R^{(gt)} \in SO(3), t^{(gt)} \in \mathbb{R}^3$ be the ground truth transformation between $X$ and $Y$ and let $R^* \in SO(3), t^* \in \mathbb{R}^3$ the estimated transformation.
The SRE between $X$ and $Y$ is defined as:
\begin{align}
\small
    SRE(X, Y) &= \frac{1}{n} \sum_{i=1}^n\frac{||R^{(gt)} x_i + t^{(gt)}) - (R^* x_i + t^*)||}{||(R^{(gt)} x_i + t^{(gt)}) - (R^{(gt)} \bar{x} + t^{(gt)})||} \\
    \bar{x} &= \frac{1}{n} \sum_{i=1}^n x_i
\end{align}

For each pair of coins from the same die, we compute the SRE.
Then, to aggregate the results for each die, we use the median instead of the mean (the mean is sensitive to outlier results and is not representative of the results, as explained in~\cite{fontana2020benchmark}).

\subsubsection{Methods}
For this benchmark, we decided to compare several methods: a classical method such as ICP\cite{Besl1992AMF} and feature matching methods with hand-crafted features such as Fast Pair Feature Histogram (FPFH~\cite{rusu_fast_2009}) and with features from deep learning methods like FCGF~\cite{Choy_2019_ICCV} and Distinctive 3D local deep descriptors (DIP~\cite{Poiesi2021}). For each die, we compute the median of SRE as explained above (reported in Table~\ref{tab:registration}), and in Table~\ref{tab:registration_av}, we report the average of SRE over the different dies in reverses and obverses.
ICP~\cite{Besl1992AMF} is very dependant on the initialization. To overcome this drawback, \cite{horache2020auto} performs random initializations. We tested using the same strategy.
FPFH~\cite{rusu_fast_2009} uses pair point feature and normals in order to compute compact local descriptors.  
FCGF~\cite{Choy_2019_ICCV} is the deep learning method we used to build the Riedones3D dataset. It uses a U-Net achitecture to compute descriptors on every point.
DIP~\cite{Poiesi2021} computes descriptors on local patches using a PointNet architecture.
Deep learning methods are said to be only effective with large training datasets, but our results are a counter-example for DIP and FCGF.
We can see in Table~\ref{tab:registration_av} that deep learning based methods outperform other methods. It shows that a small dataset is enough to learn meaningful features.
DIP and FCGF have satisfying results on obverse and reverse. However, DIP has lower results than FCGF in R8 and R5 (see Table ~\ref{tab:registration}). It shows that it is harder for DIP to generalize on unseen patterns, whereas usually DIP has better generalization capabilities. It is because in registration scenes such as ETH or 3DMatch, patterns are not as sophisticated as in Riedones3D. 

\begin{table*}[ht]
    \centering
    \small
    \begin{tabular}[t]{l|ccccc}
    \toprule
         \textbf{Methods} & \textbf{Reverses (R)} & \textbf{Obverses w beard (DB)} & \textbf{Obverses w/o beard (D)} & \textbf{All} & \textbf{Time (in s)}\\
    \midrule
Random Search ICP~\cite{horache2020auto} & 512.9 & 70.7 & 7.2 & 302.2 & 70.7\\
\midrule
FPFH (5000) + TEASER & 492.0 & 448.1 & 358.6 & 452.6 & 2.1 \\
FPFH (5000) + TEASER + ICP & 501.5 & 378.1 & 616.6 & 499.6 & 2.3 \\
DIP (5000) + TEASER & 370.4 & 46.1 & 21.9 & 220.9 & 61.2\\
DIP (5000) + TEASER + ICP & 281.4 & 73.4 & 9.3 & 174.7 & 62.9 \\
FCGF (250) + TEASER & 126.0 & 42.6 & 19.3 & 83.8 & 0.6\\
FCGF (5000) + TEASER & 96.7 & 21.4 & 10.3 & 60.8 & 2.1\\
FCGF (250) + TEASER + ICP & 101.0 & 11.0 & 9.3 & 60.6 & 1.2\\
FCGF (5000) + TEASER + ICP & 68.2 & 8.7 & 9.3 & 41.9 & 3.8\\
\bottomrule
    \end{tabular}
    \caption{SRE (x1000) on Riedones3D dataset for coin registration. For FCGF, we tried FCGF (250) and FCGF (5000): 250 and 5000 are the number of descriptors kept for the transformation estimation step. It helps to go faster in the transformation estimation. We use TEASER to estimate the transformation in a robust way. +~ICP means an additional ICP step after the registration algorithm to refine the transformation. For deep learning methods (FCGF and DIP), training is done on obverses without beard only.}
    \label{tab:registration_av}
\end{table*}

\begin{table*}[ht]
\centering
\scalebox{0.47}{
\begin{tabular}[t]{l|cccccccccccccccccccccccccccc}
\toprule
Methods & R1 & R10 & R11 & R12 & R14 & R16 & R17 & R2 & R3 & R4 & R5 & R6 & R7 & R8 & R9 & DB1a &DB2a & DB3a & DB1b & DB2b& DB3b & D1& D2 & D5 & D10 & D15 & D33 & Average\\
\midrule
Random Search ICP~\cite{horache2020auto} & 115.8&1559.7&1803.5&27.5&54.8&1.5&258.3&101.0&24.1&12.7&399.4&7.2&5.4&1595.3&1727.1&24.2&337.1&13.5&5.8&22.7&21.1&6.6&1.5&12.6&9.7&7.6&4.9& 302.2 \\
\midrule
FPFH (5000) + TEASER & 66.6&155.4&1591.2&38.2&69.1&48.3&27.9&740.8&1324.0&59.3&1504.7&39.7&23.6&1670.4&20.9&166.1&57.6&25.4&1268.7&1145.9&25.0&1233.1&548.6&103.4&56.0&180.0&30.6& 452.6\\
FPFH (5000) + TEASER + ICP & 11.2&14.4&1518.5&20.9&42.8&10.1&738.1&10.4&1321.6&21.7&1578.2&18.3&10.0&2194.6&11.2&81.4&5.4&7.5&1123.2&1043.8&7.5&1305.2&980.1&100.6&13.5&1295.7&4.6& 499.6 \\
DIP (5000) + TEASER & 29.7 &43.0 &1395.8 &32.6 &55.4 &30.6 &63.0 &28.8 &33.0 &34.3 &2027.7 &27.6 &15.4 &1705.3 &33.7 &30.5 &25.4 &23.1 &140.7 &31.1 &25.6 &22.7 &26.2 &22.7 &21.6 &30.5 &7.9 & 220.9\\
DIP (5000) + TEASER + ICP & 8.8&11.6&1546.9&24.2&47.3&9.3&24.3&10.4&11.5&9.0&595.3&10.8&9.6&1892.7&8.8&11.6&10.7&7.2&389.2&14.3&7.6&10.4&8.6&12.2&9.9&11.2&3.3 & 174.7\\
FCGF (250) + TEASER & 47.1&59.7&858.6&41.8&72.0&53.8&108.7&77.3&64.2&59.8&133.8&54.0&85.0&102.7&71.6&53.5&42.3&41.5&48.9&35.5&33.8&22.8&21.7&24.5&22.1&16.6&8.3 &  83.8 \\
FCGF (5000) + TEASER &18.5&23.4&1053.9&18.0&49.5&15.8&35.4&18.4&22.3&24.2&41.6&17.4&42.6&53.7&16.3&23.4&29.4&18.8&24.7&17.1&15.0&10.6&7.3&9.8&10.5&12.5&11.0 & 60.8\\
FCGF (250) + TEASER + ICP & 10.6&12.7&869.4&24.3&82.7&8.9&66.8&9.4&10.8&13.7&93.1&9.7&8.4&281.5&12.6&14.9&7.9&11.1&4.9&14.4&12.6&10.5&7.4&12.5&10.0&11.2&4.4 & 60.6 \\
FCGF (5000) + TEASER + ICP &9.8&9.5&731.4&20.6&61.6&10.4&32.1&10.1&9.2&8.7&70.9&8.4&9.6&22.1&8.0&10.7&6.5&9.0&4.8&12.9&8.1&10.5&8.3&11.3&9.7&11.2&5.0& 41.9\\
\bottomrule
\end{tabular}
}
\caption{Detailed results with SRE (x1000) on Riedones3D dataset for coin registration. Results are presented for each die separately.}
\label{tab:registration}
\end{table*}%

\begin{figure}
     \centering
     \begin{subfigure}[b]{\linewidth}
         \centering
         \includegraphics[width=0.5\linewidth]{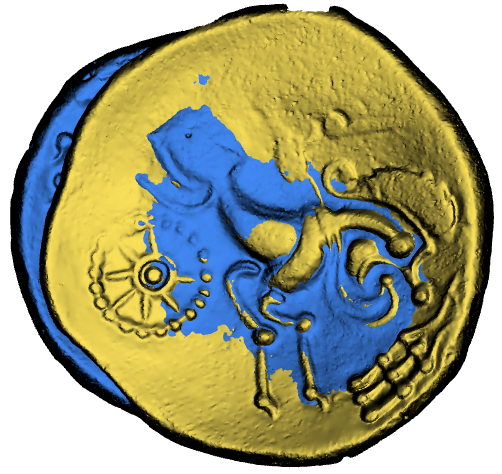}
         \caption{Example of successful registration (R5 die).}
         \label{fig:r5}
     \end{subfigure}
     \hfill
     \\
     \begin{subfigure}[b]{\linewidth}
         \centering
         \includegraphics[width=0.5\linewidth]{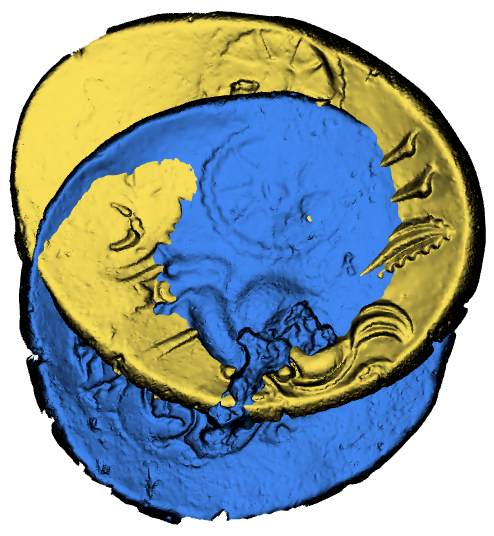}
         \caption{Example of failure for registration (R11 die).}
         \label{fig:r11}
     \end{subfigure}
        \caption{Registration on a pair of coins using FCGF\cite{Choy_2019_ICCV} on the die R5 between L0020R and L0061R (success) and a pair on the die R11 between L0053R and L0059R (failure). }
        \label{fig:qualitative}
\end{figure}
If we look at the R11 die, every method seems to fail (see Table~\ref{tab:registration}). It is because the coin L0059R has a flaw (see Figure~\ref{fig:qualitative} for the registration result and see Figure~\ref{fig:example} to see the coin L0059R). It shows that Riedones3D is a challenging dataset. Therefore, progress needs to be made to deal with defects in coins.

\subsection{Fine-grained clustering}
We present the dataset and a baseline for the die clustering on Riedones3D. For this task, we use more data for training and testing for a proper evaluation.
For the obverses, the training set has 418 coins (200 have been used for registration), the validation set has 181 coins, and the test set has 288 coins. For the reverses, the training set has 510 coins, the validation set has 299 coins, and the test set has 293 coins. For the obverse with beard, we do not have enough data, so it will only be used as a test set.

\begin{figure*}
    \centering
    \includegraphics[width=0.9\textwidth]{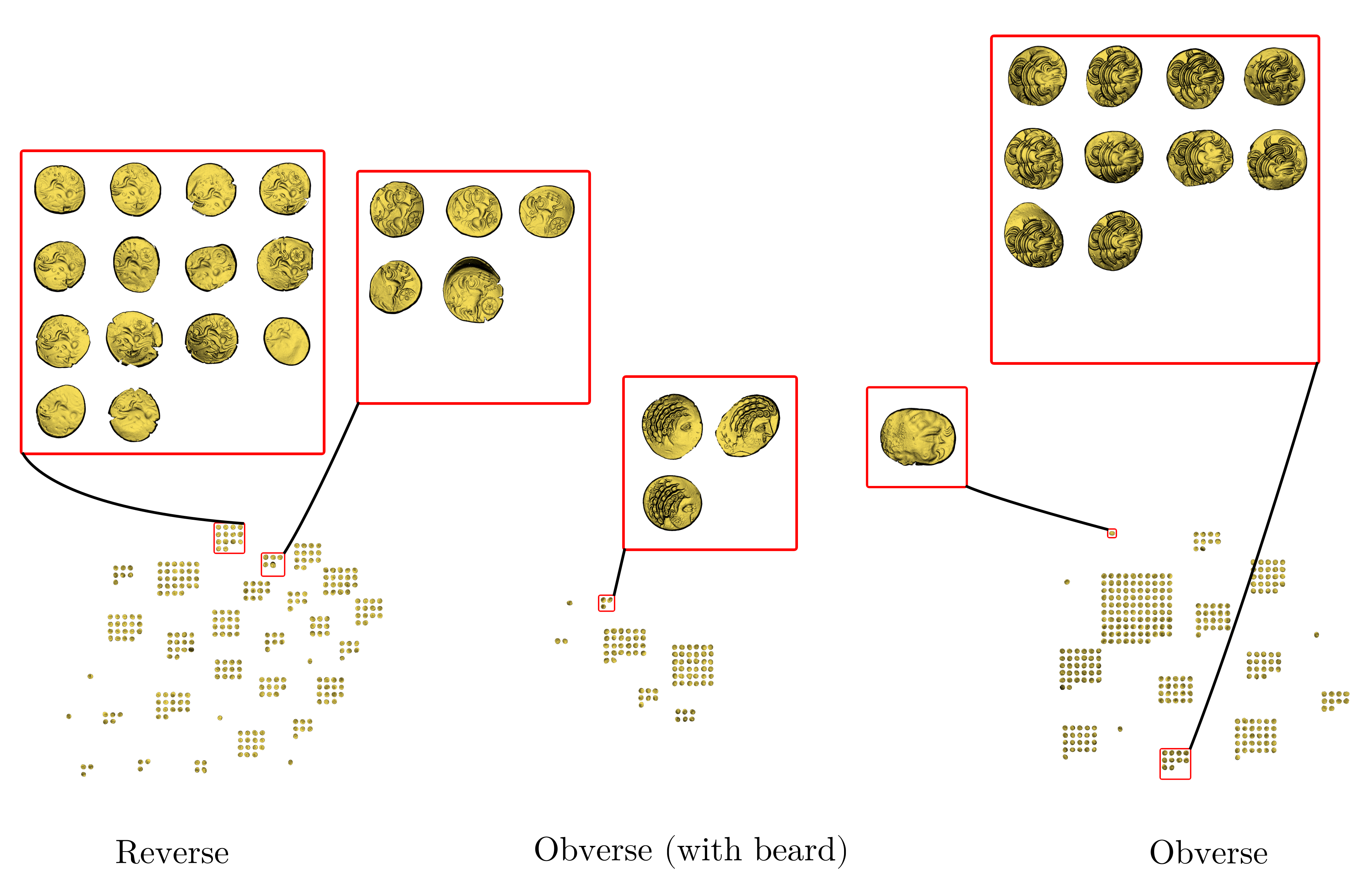}
    \caption{Synthetic view of the test set of Riedones3D for the fine-grained clustering (30 dies for reverses, 7 dies for obverses with beard and 14 dies for obverses without beard).}
    \label{fig:cluster}
\end{figure*}

Figure~\ref{fig:cluster} shows a synthetic view of the test set for die clustering. It shows that the number of coins per die is highly unbalanced. Indeed, each die does not give the same number of coin: some dies are more resistant and some are more fragile.

We propose a baseline for clustering the coins, based on the method used to build the Riedones3D dataset (registration + pairwise similarity estimation + clustering). For the registration step, we use the results from FCGF~\cite{Choy_2019_ICCV}. After the alignment, we compute the pairwise similarity from the c2c distance histogram, compute the graph, and extract the clusters with the threshold $\tau$ (as explained above for the dataset construction). On the validation set of obverses without beard, we found that the best threshold $\tau$ was 0.95.

As a metric to evaluate the clustering, we use the Fowlkes-Mallows Index (FMI)~\cite{fowlkes1983cluster}. The FMI is defined as followed:
\begin{align}
    FMI = \frac{TP}{\sqrt{(TP+FP)(TP+FN)}}
\end{align}
TP is the True Positive (the number of pairs of coins that have the same predicted die and are effectively from the same die).
TN is the True Negative (the number of pairs of coins that do not have the same predicted die and are not from the same die).
FN is the False Negative (the number of pairs of coins that have a different predicted die but are from the same die). FP is the False Positive (the number of pairs of coins that have the same predicted die but are not from the same die).

We also report the Adjusted Rand Index (ARI) \cite{Hubert1985}:
\begin{align}
\small
    RI &= \frac{TP+TN}{TP + FP + TN +FN}\\
    ARI &= \frac{RI - E[RI]}{\max(RI)-E[RI]}
\end{align}
$RI$ is the Rand Index ($TP$, $TN$, $FP$, and $FN$ are defined above). 
$E[RI]$ is the expected Rand Index for a random variable, and $\max(RI)$ is the maximum index we have.

These two metrics are classical metrics for clustering.
We can see in Table~\ref{tab:nmi} that ARI and FMI give close results on the test set of Riedones3D. The results are satisfying for the obverses, but for the reverses, there are still room for improvements. The proposed method is a good baseline for our dataset.
\begin{table}[ht]
    \centering
    \small
    \begin{tabular}{l|ccc}
    \toprule
    & Reverses & Obverses (w beard) & Obverses (w/o beard) \\
    \midrule
    FMI & 0.87 & 0.99 & 0.98 \\
    ARI & 0.86 & 0.99 & 0.97\\
    \bottomrule
    \end{tabular}
    \caption{Measure of die clustering using FMI and ARI metrics for test sets of Riedones3D (training is done on obverses without beard only).}
    \label{tab:nmi}
\end{table}
For the clustering of the obverses without beard, we obtained 16 clusters with the evaluated method, whereas there are 14 clusters in the ground truth. For the clustering of the obverses with beard, there is only one mistake: one coin not associated to the right die.
For the clustering of the reverses, we obtain 35 clusters (there are 30 clusters in the ground truth). Therefore, the algorithm could correctly identify the main groups, but still some coins are misclassified. 

\section{Conclusion}
We propose a new challenging dataset of 3D scans of coins for die recognition. We also propose a strong baseline for 3D scan registration and die clustering. 
We showed that, for registration, we can use deep learning methods. However, more investigation needs to be done in order to improve die clustering. In future research, we will investigate End-to-End deep learning methods for die clustering on Riedones3D.

\paragraph*{Acknowledgments}
This work was granted under the funding of the Idex PSL with the reference ANR-10-IDEX-0001-02 PSL. This work was granted access to the HPC resources of IDRIS under the allocation 2020-AD011012181 made by GENCI.

\bibliographystyle{eg-alpha-doi} 
\bibliography{article}       

\end{document}